\documentclass[11pt]{article}
\usepackage{acl2016}
\usepackage{times}
\usepackage{url}
\usepackage{latexsym}
\usepackage{times}
\usepackage{url}
\usepackage{graphicx}
\usepackage{color}

\aclfinalcopy 

\title{Acronym Disambiguation: A Domain Independent Approach}

\author{Aditya Thakker \textsuperscript{*} \\
  Dwarkadas J. Sanghvi  \\
  College of Engineering \\
  { aditya.thakker@djsce.edu.in} \\\And
  Suhail Barot \textsuperscript{*} \\
  Dwarkadas J. Sanghvi  \\
  College of Engineering \\
  { suhail.barot@djsce.edu.in}  \\\And
  Sudhir Bagul \\
  Dwarkadas J. Sanghvi  \\
  College of Engineering \\
  { sudhir.bagul@djsce.ac.in} \\}

\date{}

\begin{document}
\maketitle
\begin{abstract}
  Acronyms are present in usually all documents to express information that is repetitive and well known. But acronyms can be ambiguous because there can be many expansions of the same acronym. In this paper, we propose a general system for acronym disambiguations that can work on any acronym given some context information it is used in. We present methods for retrieving all the possible expansions of an acronym from Wikipedia and AcronymsFinder.com. We propose to use these expansions to collect the context in which these acronym disambiguations are used and then score them using a deep learning technique called Doc2Vec. All these things collectively lead to achieving an accuracy of 90.9\% in selecting the correct expansion for given acronym on a dataset we scraped from Wikipedia with 707 distinct acronyms and 14,876 disambiguations.
\end{abstract}

\section{Introduction}

  Acronyms are short descriptors made from important initial letters of a phrase. The phrase here is referred as an expansion of that acronym. Acronyms are used within these documents to shorten complicated or oft-repeated terms.

Acronym usage is becoming more and more common in emails, tweets, blog posts, etc. And with the increasing popularity of mobile devices, the use of acronyms on social platforms has increased even more because typing in these devices is difficult and acronyms provide a succinct way to express information. 

Usually, acronyms will be conveniently defined at the point of the first usage, but sometimes a document will omit the definition entirely, assuming the reader’s familiarity with the acronym. For example, “WHO” is often used as an acronym for “World Health Organization” and usually people are expected to know the expansion of it. Or take “CSS” as an example, most of the documents won’t even mention the expansion of “CSS” because it’s such a common acronym for “Cascading Style Sheets”. But “CSS” can also mean “Content-Scrambling System”, “Closed Source Software”, and “Cross-Site Scripting”. 

Also, many natural language processing applications require preprocessing of a document. Text normalization is one of the most important phase of these preprocessing tasks. The basic task of text normalization is to convert non-standard words like numbers, abbreviations, dates, etc. into standard words, though depending on the task and the domain a greater or lesser number of these non-standard words may need to be normalized. In this phase of text normalization, we need to expand all the acronyms in the document. Acronyms are typically ambiguous because several expansions exist for the same acronym as we saw in the example before. For example, “Cable News Network” and “Convolutional Neural Network” are both expansions for the common acronym “CNN”. To disambiguate these acronyms, we can use context paragraphs that surround these acronyms to find the actual expansion. 
\footnote{ \textsuperscript{*} indicates these author is an equal contributor to this work}
\begin{figure*}[t]
\centering
    \includegraphics[width=14cm,height=7cm]{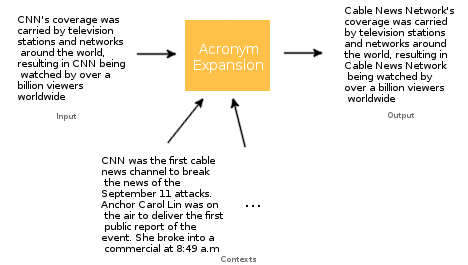}
    \caption{Acronym Exoansion from Input to Output}
    \label{fig:methodology}
\end{figure*}

In our work, we have studied and created an information retrieval system which takes any acronyms along with some context words and then will expand the acronym based on the score it gives to all the possible expansions on the acronym. As shown in the figure, the system will search for all the possible expansions of the given acronym on Wikipedia and Acronymfinder.com. Once it has the list of all the expansions then it will start finding occurrences of those phrases in Wikipedia to get all the contexts in which it is used. Our system will then represent each possbile expansion using a deep learning technique called Doc2Vec \cite{Mikolov:2014} in high dimensional vector space. Doc2Vec \cite{Mikolov:2014} which is used in our system can be seen as a distributional semantic representation and this representation is proved to be effective to compute the semantic similarity between words based on the context without any labeled data. The Doc2Vec \cite{Mikolov:2014} embeddings represents the expansions of acronyms in vector space. The placement of each acronym expansion depends on the context that it is used in. Once the system has represented all the possible context vectors associated with each expansion using Doc2Vec \cite{Mikolov:2014}, we can pick the expansion whose context vector has the highest cosine similarity score with the input context vector which will then be our expansion for that given acronym.

To the best of our knowledge, we are the first to apply Doc2Vec \cite{Mikolov:2014} embeddings to this task. Experimental results show that our system achieves a comparable accuracy of 90.9\% accuracy and is close to human’s performance. 

Our paper is mainly divided into the following sections:
\begin{itemize}
\item In Section 1, we begin with an introduction to the task of acronym expansion and briefly describe our approach. 
\item In Section 2, we mention the issues with acronym expansion and provide an overview of the  past approaches to the same problem.
\item In Section 3, we descibe our proposed approach to the task of acronym expansion and the creation of document embeddings from context of acronym usage which is at the core of our model.
\item In Section 4, we explain our experimental setup, describe how we gathered the dataset  and give results and observations of testing on the datasets.
\item In Section 5, we give our conclusions from the experiments and also describe methods to extend our approach to similar problems.
\end{itemize}

\section{Related Work}   

The task of acronym expansion has been intensively studied by various researchers using supervised learning algorithms. However, the performance of these supervised methods depends on a large amount of labeled data which is extremely difficult to obtain. 

In Hippocratic Abbreviation Expansion \cite{Roark:2014} paper, they have used SVM, N-Gram, and many hand-crafted feature engineering techniques to identify the correct expansion of an acronym.

In Acronym-Expansion Recognition and Ranking on the Web \cite{Jain:2007}, they use a very similar technique of information retrieval to find all the expansions of any acronyms and then ranked them using co-occurence between acronym and expansion, popularity and reliability of sources. 
 
One other difference between the work we report from much of the recent work cited above is that our work focuses on a more general system to solve the problem. Most of the recent works we have mentioned before are focused on some particular domain and hence use some domain specific techniques to achieve better accuracy. Our system on the other hand only uses the textual data present on Wikipedia to understand the context and outputs the closest expansion similar to input context.  

\section{Proposed Approach}

Owing to the recent success in deep learning frameworks, we sought to apply the techniques to Acronym Expansion problem. But, the main challenge in these approaches is to identify the correct expansion inspite of the many expansions for the same acronym. 

We propose to use the vast amount of data available on the internet to identify the correct expansion for any acronym. Our approach involves using Document embeddings to understand the context in which an acronym is used. Document embeddings \cite{Mikolov:2014} are a direct extrapolation of the concept of Word Embeddings \cite{Mikolov:2013}. We extract the paragraphs where the acronym was used and supply it to our model. These paragraphs are then embedded in high dimensional vector space, where vector proximity is a direct measure of similarity of context. This concept is explained further in detail in the following sections.

\begin{figure}[h]
\centering
    \includegraphics[width=4cm]{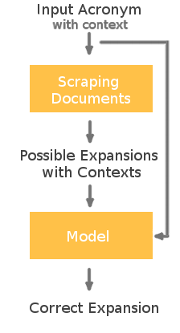}
    \caption{Our Approach}
    \label{fig:our_approach}
\end{figure}

\subsection{Crawling Data}
As shown in Figure 3, an acronym is given to our system as input. The input is then used to search for all the expansions that we can find for it. To identify, whether any phrase is an expansion of the given acronym, we have made 3 conditions that it must follow: 
\begin{itemize}
\item The first letters of the words must match the acronym on the sequence
\item The words can be separated using space ( ), underscore(\_) or dash (-)
\item It can consist of stop words in between if the first letters do not match
\end{itemize}

Implementing these rules, we were able to crawl almost all the expansions that are possible of an acronym.

\begin{figure*}[t]
\centering
    \includegraphics[width=11cm]{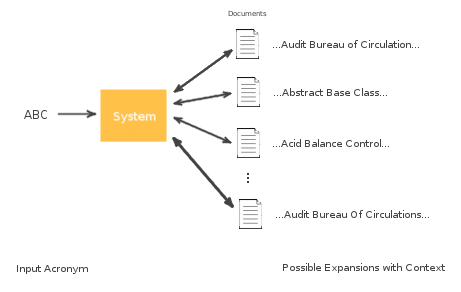}
    \caption{Crawling and finding possible expansions}
    \label{fig:crawling}
\end{figure*}

After finding all the expansions that we could crawl, we had a list of expansions that were possible expansions for the given input. Now, to find the correct expansion, we wanted some contextual data that was used when these expansions were mentioned in any document. Our system would then use the list of expansion to further search for all the occurrences of that expansion and collected some data that surrounds it. This surrounding data is the contextual data that we need to identify the correct expansion of the acronym given to us. The amount of data (words) that we picked surrouding the expansion was of size ranging from 2000-5000 characters (at max). By 2000, we mean that words in 1000 characters before the expansions and words in 1000 characters after the expansion. 

It might happen that our system would select an expansion-context pair even if the same expansion has already been fetched. We have purposely allowed it because even if the expansion is same, the context will be different in which the acronym is used. The different contexts for same expansion helps the system to find the correct expansion.

\subsection{Model}
It has become common practice to use word embeddings \cite{Mikolov:2013} for semantic analysis, the most famous implementations being Google’s Word2Vec \cite{Mikolov:2013} and Stanford’s GloVe \cite{Pennington:2014}. However, researchers have been experimenting, with great success, with sentence/paragraph/document embeddings - commonly known as thought vectors -  for the past few years. Our model is based on Google’s Doc2vec \cite{Mikolov:2014} model. It is a neural network architecture that outputs N (number of paragraphs) labelled vectors each of  M dimensions.

We have trained our datasets on both the models proposed by Doc2Vec \cite{Mikolov:2014}, namely the distributed memory model and distributed bag of words model. 
The distributed memory model takes into account the context of the surrounding words while predicting a word, while the distributed bag of words model does not.

According to Doc2Vec \cite{Mikolov:2014}, given a set of training words, we maximise average log likelihood as:

\[
\frac{1}{T}\sum_{t=k}^{T-k} \log p (w_t | w_{t-k}, ..., w_{t+k})
\]

As per their model, prediction is handled by a multiclass classifier (softmax) :

\[
p (w_t | w_{t-k}, ..., w_{t+k}) = \frac{e^{y_{w_{t}}}}{\sum_i{e^{y_i}}}
\]

We got marginally better results on the Doc2Vec model, as compared to the Distributed Bag of Words model.

As mentioned earlier, our implementation was based on the Doc2vec model.

\section{Experiments}

\begin{table*}[t]
\centering
\resizebox{\linewidth}{!}{%
\begin{tabular}{|l|l|l|l|l|l|}
\hline \bf Doc2Vec Model & \bf Embedding Size. & \bf Context/Source & \bf Length of Source/Context & \bf Traning Epochs & \bf Accuracy\\ \hline
Distributed Bag of Words & 500 & Context & - & 12 & 88.9\% \\
Distributed Bag of Words & 500 & Context & - & 12 & 89.7\% \\
Distributed Bag of Words & 500 & Context & 2000 & 12 & 90.7\% \\
Distributed Bag of Words & 500 & Context & 2000 & 12 & 90.6\% \\
Distributed Bag of Words & 200 & Source & 2000 & 12 & \textbf{90.9\%} \\
Distributed Memory & 200 & Context & 5000 & 12 & 88.4\% \\
Distributed Memory & 750 & Context & 2000 & 15 & 89.7\% \\
Distributed Memory & 200 & Source & 5000 & 15 & 86.1\% \\
Distributed Memory & 500 & Context & 5000 & 15 & \textbf{90.9\%} \\
\hline
\end{tabular}}
\caption{Results of experiments}
\end{table*}

We wanted to be absolutely comprehensive in our approach, so we scraped 707 distinct acronyms with their occurences and context in which they had occured. 

We got marginally better results on the distributed memory model, as compared to the distributed bag of words model.

For each acronym, we train a model with all the context possibilities. We then calculate the cosine similarity between every input-context and crawled-context pair. Following that, we extract the pair with the highest cosine similarity value. To give some physical intuition, this means that this pair of vectors are the closest together in vector space.
We predict that the full form associated with the context selected above is the same as the full form associated with the meaning. Using python’s in built sequence matcher, we match the predicted expansion with the expansion associated with the input context to verify the models prediction and calculate accuracy. 

So, for example, if CNN is the acronym at hand, we have one context paragraph and a expansion (Convolutional Neural Network) associated with it, and several crawled context paragraphs (i.e. places on Wikipedia articles where the acronym CNN has occurred). Each context paragraph also has a distinct expansion associated with it. Let’s take two distinct context paragraphs, one with a expansion of "Cable News Network" associated with it, and another with the expansion "Convolutional Neural Network" associated with it. 

We plot all 3 paragraphs in vector space, and calculate the cosine similarity of the input context and all the crawled-contexts pair-wise. So here, cos\_sim(input\_context, crawled\_context\_1) and cos\_sim(input\_context, crawled\_context\_2) are compared. Now we select the pair with the highest cosine similarity, let’s say, (input\_context, crawled\_context). meaning has a full form of "Convolutional Neural Network" associated with it. If crawled\_context also has a full form of "Convolutional Neural Network" associated with it, then our model has worked successfully, otherwise not.

\begin{figure}[h]
\centering
    \includegraphics[width=\linewidth]{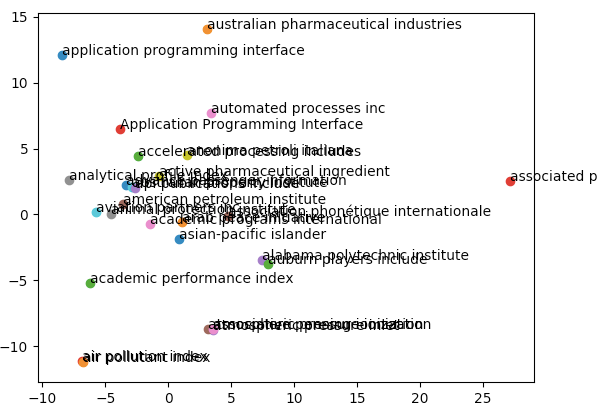}
    \caption{Doc2Vec \cite{Mikolov:2014} Plot for Acronym 'API'}
    \label{fig:plot}
\end{figure}

The Figure 4 is an approximate plot of the vector space for the acronym API. This was achieved using Principal Component Analysis. Keeping in mind that 500 dimensions are being condensed to 2 dimensions, this plot is for representation purposes only, and is in no way indicative of the model’s accuracy.

Using the dataset mentioned before, we ran our model on a total of  14,876 disambiguations for 707 distinct acronyms. We achieved an accuracy of 90.9\%.

\subsection{Experimental Setup}

We use this architecture for the network because of the constraint on the dataset size caused by scarcity of labelled data. We used a NVIDIA 970 GTX GPU and a  4.00 GHz Intel i7-4790 processor with 64GB RAM to train our models. As the datasets in this domain expand, we would like to scale up our approach to bigger architectures. The results obtained on different experiments are given in Table 1. We are able to achieve comparable accuracies without using any domain specific feature engineering.

\subsection{Observations}

A crawled input for our model ranges from 200 characters to 60,000 characters, as we wanted to simulate real life scenarios as much as possible. 
A learning rate of 0.025 was found to be ideal, coupled with 12 epochs of training the same model. 
Less than 10 epochs proved to cause a significant decrease in accuracy due to undertraining.
Greater than than 15 epochs of training caused the same problem, but due to overtraining,
vectors of 500 dimensions for Distributed Memory model and vectors of 200 dimensions for Distributed Bag of Words model proved to be ideal on our datasets. On smaller paragraphs, smaller dimensions of vectors (100-150) seemed to lead to more accurate predictions, whereas on larger prargraphs, larger dimension vectors(800-1000) worked better.

In some special cases, if an acronym is found in contexts with other acronyms, the models accuracy decreases. For example, in case of acronym "ETC", it can found in context of "European Travel Commission" also. So the cosine similarity score of "European Travel Commission" will be very close to that of "Et Cetera". 
\footnote{Code available at: https://github.com/adityathakker/AcronymExpansion}

\section{Conclusion}

The experimental results have shown that document embeddings are a promising solution to the acronym disambiguation problem. The results we achieved are stable even without using any hand-crafted feature engineering which proves that it's a general data-oriented system.

For further work, we want to try this approach to make recommendation engines that use such contextual data that surrounds any (product) name to identify similar (product) names and recommend them to users.


\begin{thebibliography}{}

\bibitem[\protect\citename{Mikolov \bgroup et al.\egroup }2014]{Mikolov:2014}
Tomas Mikolov, Quoc V. Le.
\newblock 2014.
\newblock Distributed Representations of Sentences and Documents.
\newblock {\em Proceedings of the 31 st International Conference on Machine Learning, Beijing, China},.

\bibitem[\protect\citename{Roark \bgroup et al.\egroup }2014]{Roark:2014}
Brian Roark, Richard Sproat.
\newblock 2014.
\newblock Hippocratic Abbreviation Expansion.
\newblock {\em Proceedings of ACL 2014},.

\bibitem[\protect\citename{Jain \bgroup et al.\egroup }2007]{Jain:2007}
Alpa Jain and Silviu Cucerzan and Saliha Azzam
\newblock 2007.
\newblock Acronym-Expansion Recognition and Ranking on the Web.
\newblock {\em IEEE International Conference on Information Reuse and Integration (2007)},.

\bibitem[\protect\citename{Collobert \bgroup et al.\egroup }2011]{Collobert:2011}
Ronan Collobert,Jason Weston,Leon Bottou,Michael Karlen,Koray Kavukcuoglu,Pavel Kuksa.
\newblock 2011.
\newblock Natural Language Processing (almost) from Scratch.
\newblock {\em Journal of Machine Learning Research, 12:2493-2537, 2011},.


\bibitem[\protect\citename{Lin \bgroup et al.\egroup }2009]{Lin:2009}
Dekang Lin and Xiaoyun Wu.
\newblock 2009.
\newblock Phrase clustering for discriminative learning..
\newblock {\em Proceedings of the Joint Conference of the 47th Annual Meeting of the ACL and the 4th International Joint Conference on Natural Language Processing of the AFNLP:},.

\bibitem[\protect\citename{Srivastava \bgroup et al.\egroup }2014]{Srivastava:2014}
Nitish Srivastava, Geoffrey Hinton, Alex Krizhevsky, Ilya Sutskever, Ruslan Salakhutdinov.
\newblock 2014.
\newblock Dropout: A Simple Way to Prevent Neural Networks from Overfitting.
\newblock {\em Journal of Machine Learning Research},.


\bibitem[\protect\citename{Mikolov \bgroup et al.\egroup }2013]{Mikolov:2013}
Tomas Mikolov, Ilya Sutskever, Kai Chen, Greg Corrado, Jeffrey Dean.
\newblock 2013.
\newblock Distributed Representations of Words and Phrases and their Compositionality.
\newblock {\em Proceedings of Neural Information Processing Systems},.

\bibitem[\protect\citename{Pennington \bgroup et al.\egroup }2014]{Pennington:2014}
Jeffrey Pennington, Richard Socher, Christopher D. Mannin.
\newblock 2014.
\newblock GloVe: Global Vectors for Word Representation.
\newblock {\em Empirical Methods in Natural Language Processing (EMNLP)},.

\bibitem[\protect\citename{Hochreiter \bgroup et al.\egroup }1997]{Hochreiter:1997}
Sepp Hochreiter , Jürgen Schmidhuber.
\newblock 1997.
\newblock Long Short-Term Memory.
\newblock {\em Journal     Neural Computation archive Volume 9 Issue 8, November 15, 1997     },.

\bibitem[\protect\citename{Kingma \bgroup et al.\egroup }2015]{Kingma:2015}
Diederik Kingma, Jimmy Ba.
\newblock 2015.
\newblock Adam: A Method for Stochastic Optimization.
\newblock {\em  International Conference for Learning Representations},.


\bibitem[\protect\citename{Sutskever \bgroup et al.\egroup }2013]{Sutskever:2013}
Ilya Sutskever,James Martens, George Dahl,Geoffrey Hinton.
\newblock 2013.
\newblock On the importance of initialization and momentum in deep learning.
\newblock {\em Journal of Machine Learning Research},


\bibitem[\protect\citename{Schuster \bgroup et al.\egroup }1997]{ Schuster :1997}
Mike Schuster and Kuldip K. Paliwal.
\newblock 1997.
\newblock Bidirectional Recurrent Neural Networks.
\newblock {\em IEEE TRANSACTIONS ON SIGNAL PROCESSING, VOL. 45, NO. 11, NOVEMBER 1997},.

\bibitem[\protect\citename{Bengio \bgroup et al.\egroup }1994]{Bengio:1994}
Yoshua Bengio, Patrice Simard, Paolo Frasconi
\newblock 1994.
\newblock Learning Long-Term Dependencies with Gradient Descent is difficult.
\newblock {\em IEEE Transactions on Neural Networks },


\end{thebibliography}
\end{document}